%% file: main.tex
\documentclass[letterpaper, 10 pt, conference]{ieeeconf}  
\IEEEoverridecommandlockouts
\usepackage{cite}
\usepackage{amsmath,amssymb,amsfonts, bm}
\usepackage{algorithmic}
\usepackage{graphicx}
\usepackage{textcomp}
\usepackage{multirow}
\usepackage{tabularx,booktabs}
\usepackage[hidelinks]{hyperref}
\usepackage{xcolor}
\usepackage{subcaption}
\usepackage{tikz}
\usepackage{pgfplots}

\usepackage{fancyhdr}
\fancypagestyle{withfooter}{
  
  \fancyfoot[C]{\footnotesize Accepted to the IEEE ICRA Workshop on Field Robotics 2024}
}

\pgfplotsset{compat=1.18}
\usetikzlibrary{spy,calc}
\usepackage{chngpage}

\def\BibTeX{{\rm B\kern-.05em{\sc i\kern-.025em b}\kern-.08em
    T\kern-.1667em\lower.7ex\hbox{E}\kern-.125emX}}

\begin{document}

\title{\LARGE \bf
    OtterROS: Picking and Programming an Uncrewed Surface Vessel for Experimental Field Robotics Research with ROS 2*
}

\author{Thomas M.~C.~Sears$^{1}$, M.~Riley Cooper$^{1}$, Sabrina R.~Button$^{1}$, and Joshua A.~Marshall$^{1}$
    \thanks{*This project was funded in part by the NSERC Canadian Robotics Network (NCRN) under NSERC project NETGP 508451-17, the Vanier Canada Graduate Scholarships program, and Queen's University's USSRF program. The Otter USV was purchased with generous support from the Ingenuity Labs Research Institute's Research Equipment Fund.}
    \thanks{$^{1}$T.~M.~C.~Sears, M.~R.~Cooper, S.~R.~Button, and J.~A.~Marshall are all with the Department of Electrical \& Computer Engineering and the Ingenuity Labs Research Institute, Queen's University at Kingston, Ontario, Canada. {\tt \{thomas.sears,joshua.marshall\}@queensu.ca}}
    \thanks{The authors do not work for, consult, own shares in or receive funding from any company or organisation that would benefit from this article, and have disclosed no relevant affiliations beyond their academic appointments.}
}

\maketitle
\thispagestyle{withfooter}
\pagestyle{withfooter}

\begin{abstract}
    There exist a wide range of options for field robotics research using ground and aerial mobile robots, but there are comparatively few robust and research-ready uncrewed surface vessels (USVs). This workshop paper starts with a snapshot of USVs currently available to the research community and then describes ``OtterROS'', an open source ROS 2 solution for the Otter USV. Field experiments using OtterROS are described, which highlight the utility of the Otter USV and the benefits of using ROS 2 in aquatic robotics research. For those interested in USV research, the paper details recommended hardware to run OtterROS and includes an example ROS 2 package using OtterROS, removing unnecessary non-recurring engineering from field robotics research activities.

\end{abstract}

\setcounter{footnote}{1}

\section{Introduction}

Aquatic mobile robots have the potential to become vital tools for environmental monitoring, infrastructure assessment, emergency response, shipping and transportation. Unfortunately, researchers with an interest in uncrewed surface vessel (USV) autonomy have been limited by the number of available platforms that are suitable for research purposes. Environmental conditions such as waves and current, affordability, and software compatibility further limit the selection of USVs for researchers seeking to field test their designs. This paper presents our group's effort to take one commercial product---the Otter USV by Maritime Robotics \cite{mr-otter-usv}---and convert it to a system that is suitable for conducting field robotics research using Robot Operating System (ROS) 2.

Offroad Robotics is a field and mobile robotics research group located on the North-East shore of Lake Ontario (Kingston, Ontario, Canada). Motivated to explore robotics in our local waterways, we decided to procure a USV. Our initial choice was the Clearpath Robotics Heron USV because it was well known and used by the robotics research community; but the Heron was discontinued before we could initiate a purchase. As a result, we conducted a survey of commercially available USVs to identify those capable of open water operations and with suitable software interfaces. Ultimately, the Otter was selected because it exceeded the Heron in every capacity, although it lacked ROS support.

\begin{figure}[t!]
    \centering
    \includegraphics[width=\columnwidth]{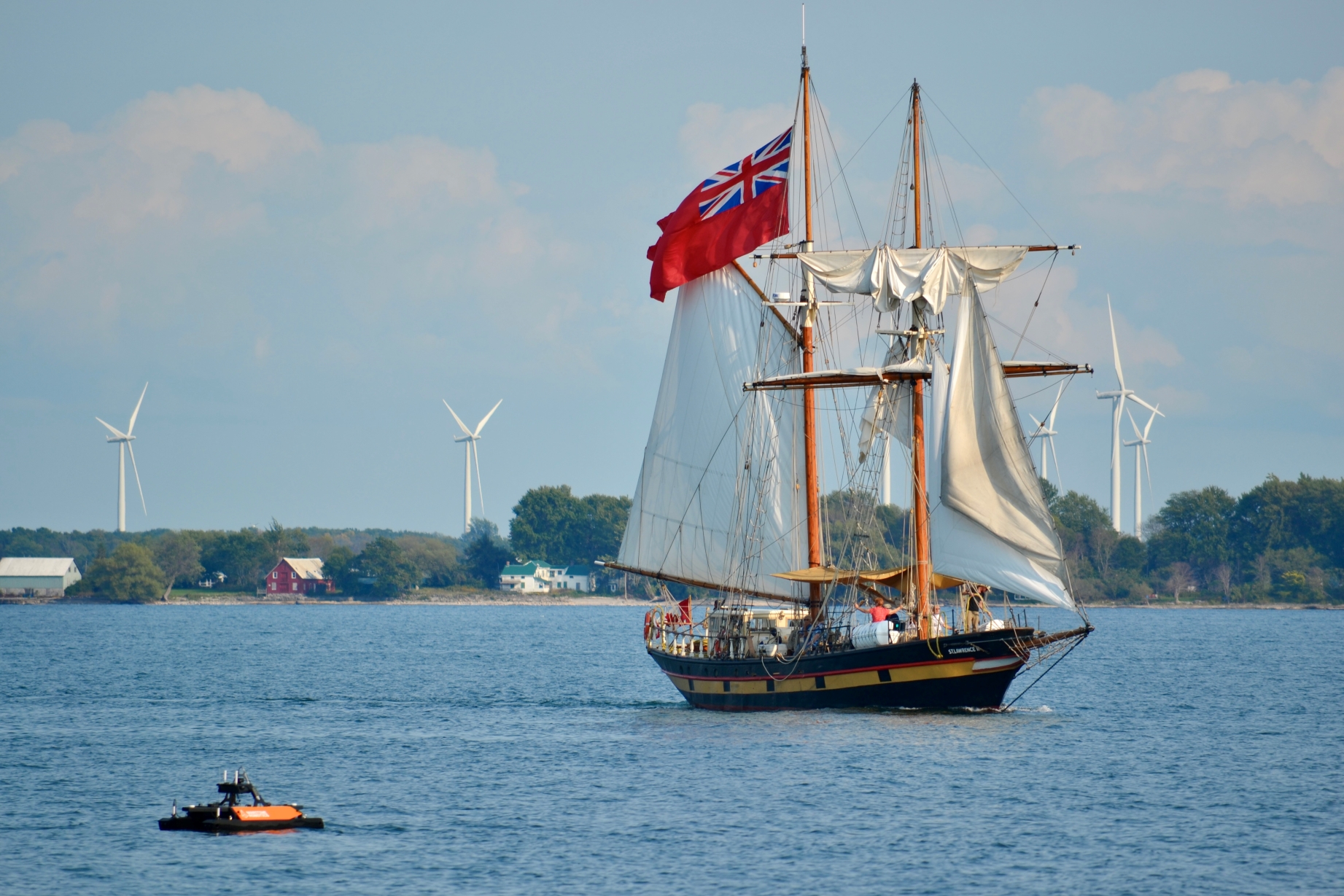}
    \caption{The Otter USV (left) during experiments in Lake Ontario. Encounters with pleasure craft are common, but occasionally we cross paths with history.} 
    \label{fig:otter_in_action}
\end{figure}

In this paper, we share hardware modifications, software developments, and lessons learned in field testing with the Otter USV over the last two years, in an effort to lower the barrier of entry to USV research for others in the field robotics community. This includes discussions about:
\begin{itemize}
    \item why the Otter was selected and how it has performed as an experimental robotics platform;
    \item the architecture and use of OtterROS, a new ROS 2 package that can interface with the Otter;
    \item hardware necessary to enable ROS on the Otter; and
    \item a control package built on OtterROS, which serves as an example of how the Otter USV and OtterROS can be used to conduct research in real environments.
\end{itemize}
This invitation to use OtterROS also includes the opportunity to share developments and algorithms that further advance autonomy for the Otter USV. Full details and source code for OtterROS can be found at: {\small\url{https://github.com/offroad-robotics/otter_ros}}.

\section{Background}

In 2022, our research team chose to purchase an Otter USV after surveying available surface vessels for field robotics research. ROS 2 compatibility was sought, but it was uncommon amongst the identified USVs and unavailable on the Otter. Table~\ref{tab:usv_survey} highlights the findings of this USV survey. This section discusses what to consider when selecting a USV in research and why we selected the Otter for our work.

\begin{table*}
    \centering
    \caption{Survey results for USVs capable of operating in open water. They vary in size, design, price, and performance. Software interfaces are varied and may require manufacturer support to enable robotics work. Configuration includes steering (S), propulsion (P), and hull (H) design. A ``--'' indicates information unavailable at time of writing.}
    \label{tab:usv_survey}
    \input{usv-survey}
\end{table*}

\begin{figure*}
    \centering
    \input{usv-survey-figure}
    \caption{Images of the USVs identified in the survey. Note: Images are not at a constant scale; see Table~\ref{tab:usv_survey} for dimensions.}
\end{figure*}

\subsection{USVs for Field Robotics}

What defines a \textit{useful} USV of course varies by user. Our group considered mechanical, electrical, and administrative factors that would affect our ability to use a USV for field work. We evaluated each of the USVs listed in Table~\ref{tab:usv_survey} against the criteria in Table~\ref{tab:usv_factors}. These factors can guide other groups in acquiring a USV suitable for their robotics work.

\begin{table*}
    \centering
    \caption{Factors to consider in USV selection for robotics research and why we at Offroad Robotics selected the Otter USV.}
    \label{tab:usv_factors}
    \input{usv_factors}
\end{table*}
\footnotetext{Source: A combination of lived experience and data from {\small\url{https://weather.gc.ca/marine/index_e.html}}}

Table~\ref{tab:usv_factors} also highlights our group's specific considerations and how the Otter addressed each factor. Multiple USVs met our needs on each individual factor, but the Otter was the best overall match. The portability, runtime, and payload capacity of the Otter were of particular interest.

\subsection{Middleware for USV Autonomy}

As is similar across many domains of robotics, multiple software libraries are publicly available and used for USV guidance, navigation, and control. MOOS, developed at the Oxford Mobile Robotics Group (MRG), is a cross-platform robotics middleware originally developed for maritime applications \cite{moos}. MIT's Laboratory for Autonomous Marine Sensing Systems (LAMSS) manages MOOS-IvP, which adds many autonomy features on core MOOS~\cite{moos}. The Underwater Systems and Technology Laboratory at the University of Porto, Portugal, created DUNE (Unified Navigation Environment), for low-level control of USVs \cite{dune}. Neptus, a ground control system with a graphical interface, allows operators to control vehicles that run DUNE. MOOS and DUNE are two options among many with origins in USV and uncrewed underwater vehicle (UUV) research, and have been deployed in numerous published works, including \cite{fonseca2021algal}, \cite{paine23}.

Since 2009, the Robot Operating System (ROS) has grown to become the de facto software development and research interface for robots~\cite{ros1}. With the latest iteration of ROS 2, the ease of implementation and flexibility of ROS has been matched with significant performance enhancements suitable for the challenges of real-world deployment~\cite{ros2}. ROS 2 is now widespread in ground and aerial mobile robotics.

Although our group was predisposed to choosing ROS 2, it is worth comparing these three main options before committing significant development effort. Other groups looking to test their own guidance, navigation, and control algorithms may consider the elements of Table~\ref{tab:middlewares} in selecting their USV software architecture. Despite ROS 2 being uncommon in USV research, Offroad Robotics found that its general popularity, available documentation, and ease of development made it the most desirable for academic research work.

\begin{table}
    \def\tabularxcolumn#1{m{#1}}
    \centering
    \caption{Comparison of major open source USV control software.}
    \label{tab:middlewares}
    \begin{tabularx}{\columnwidth}{
        >{\raggedright\arraybackslash}X
        >{\raggedright\arraybackslash}m{1.3cm}
        >{\raggedright\arraybackslash}m{1.5cm}
        >{\raggedright\arraybackslash}m{1.8cm}
        }
        \toprule
        \textbf{}                                       & \textbf{DUNE}                                                                                                                                                           & \textbf{MOOS-IvP}                                   & \textbf{ROS 2} \\
        \midrule
        First release                                   &
        2007                                            & 2001                                                                                                                                                                    & 2017                                                                 \\[0.2cm]
        Maintainer                                      &
        \href{https://www.lsts.pt/toolchain/dune}{LSTS} & \href{https://www.robots.ox.ac.uk/~mobile/MOOS/wiki/pmwiki.php/Main/HomePage}{MRG} and \href{https://oceanai.mit.edu/moos-ivp/pmwiki/pmwiki.php?n=Main.HomePage}{LAMSS} & \href{https://www.openrobotics.org/}{Open Robotics}                  \\[0.2cm]
        Development origins                             &
        USVs                                            & USVs                                                                                                                                                                    & Generic                                                              \\    [0.2cm]
        Published use\footnotemark                      &
        \raisebox{-0.8ex}{\textasciitilde}500           &
        \raisebox{-0.8ex}{\textasciitilde}1,000         &
        \raisebox{-0.8ex}{\textasciitilde}30,000 (\raisebox{-0.8ex}{\textasciitilde}100 with ``USV'')                                                                                                                                                                                                    \\[0.2cm]
        Documentation                                   &
        Limited                                         & MIT User Guide                                                                                                                                                          & ROS 2 Docs, YouTube, ...                                             \\
        \bottomrule
    \end{tabularx}
\end{table}
\footnotetext{A snapshot of search result count from March 2024 from Google scholar and GitHub for ``DUNE Unified Navigation Environment'', ``MOOS-IvP'', and ``ROS 2'' (and ``ROS 2 USV'').}

\subsection{Data Interface with the Otter USV}

Many USV manufacturers provide some form of interface with their onboard computer (OBC), and often oblige the ``backseat driver'' philosophy. This was popularized by MOOS, where low-level control and vehicle autonomy are separated~\cite{moos-jfr}. In this perspective \textit{control} is the domain of manufacturers and aims to manage the vehicle's actuators to achieve certain heading, speed, or other system states. The way the vehicle achieves these states is not necessarily of interest to the operator. \textit{Autonomy}, or mission planning, is the realm of the backseat driver system, which uses high level navigation data to decide what values to send back to the control system.

MOOS was intended for mission planning, so the backseat driver paradigm often does not provide access to the low-level telemetry or actuator control. Furthermore, this architecture does not prescribe what information should be provided to an external computer. Our recommendation is to speak with manufacturers to fully understand what can and cannot be accessed. This greatly impacts which USVs are useful for field robotics---navigation and planning experiments may only require high-level telemetry from the USV, while control research may demand direct access to motor commands.

\begin{figure}[tb]
    \centering
    \includegraphics[]{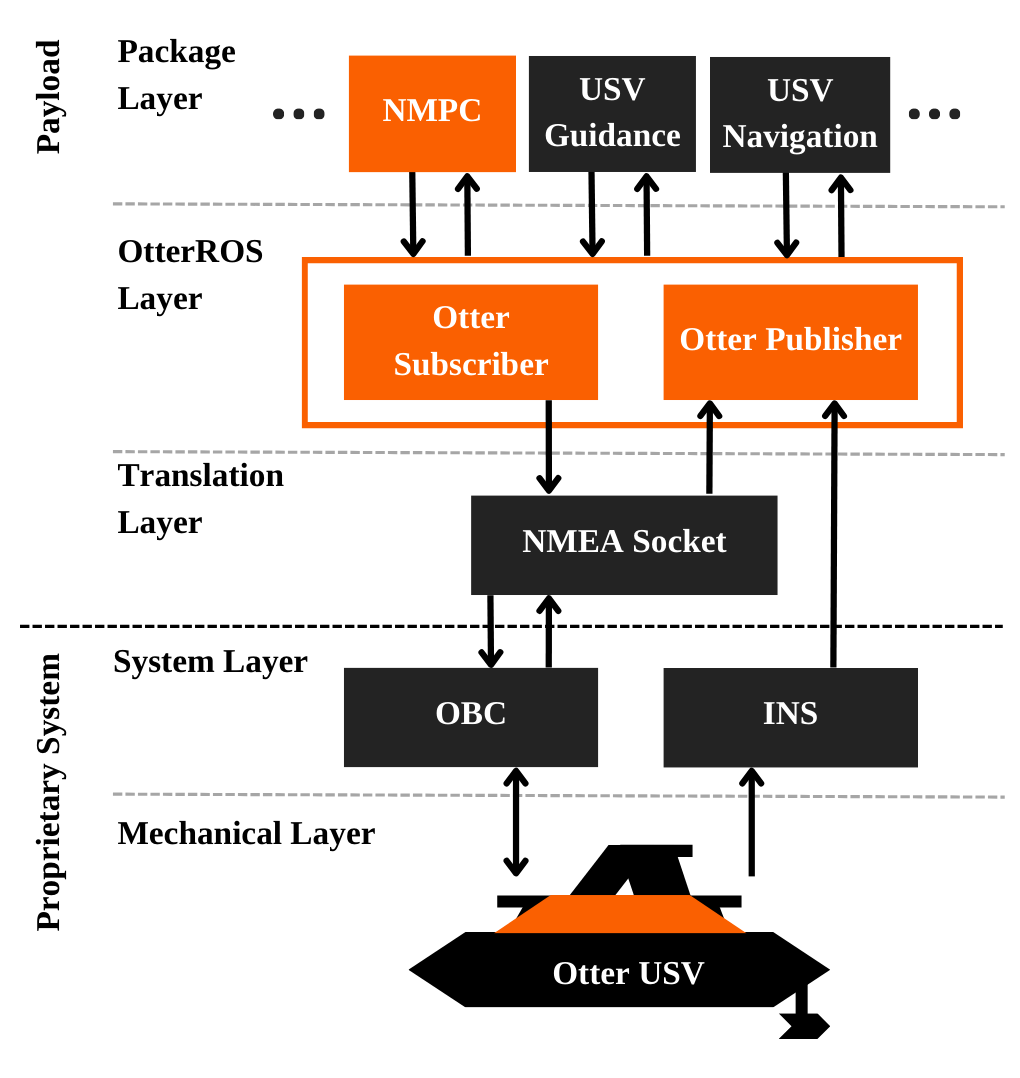}
    \caption{Adapting the Otter's external interface for ROS 2 with \textit{OtterROS}. The Payload computer running ROS 2 communicates with the Otter over a local network connection. OtterROS manages the interface between the USV and new programs in the Package Layer.}
    \label{fig:otterros-layers}
\end{figure}

Maritime Robotics provides a backseat driver interface through the Otter's onboard local area network. The OBC broadcasts navigation and system state information, including position, orientation, and speed, which any device can receive. The OBC  also listens for incoming commands for external control\footnote{This is an optional add-on from Maritime Robotics and must be enabled.}. These commands range from path following to motor thrust control, which is particularly valuable for control research. The backseat driver architecture is illustrated as the communication between the Translation and System layers of Fig.~\ref{fig:otterros-layers}. In this diagram, the ``Payload'' computer is separate from the OBC, which in our case runs ROS 2 through our package OtterROS.

\section{The Otter USV for Robotics Research}

The Otter USV is marketed towards ``efficient and precise data acquisition, environmental monitoring, and surveillance in sheltered, coastal and shallow areas''~\cite{mr-otter-usv}, which is a good starting point for a field research robot. This section highlights how the physical configuration of the Otter and the data interfaces available are suitable for robotics research.

\begin{figure*}[t!]
    \centering
    \includegraphics[width=\textwidth]{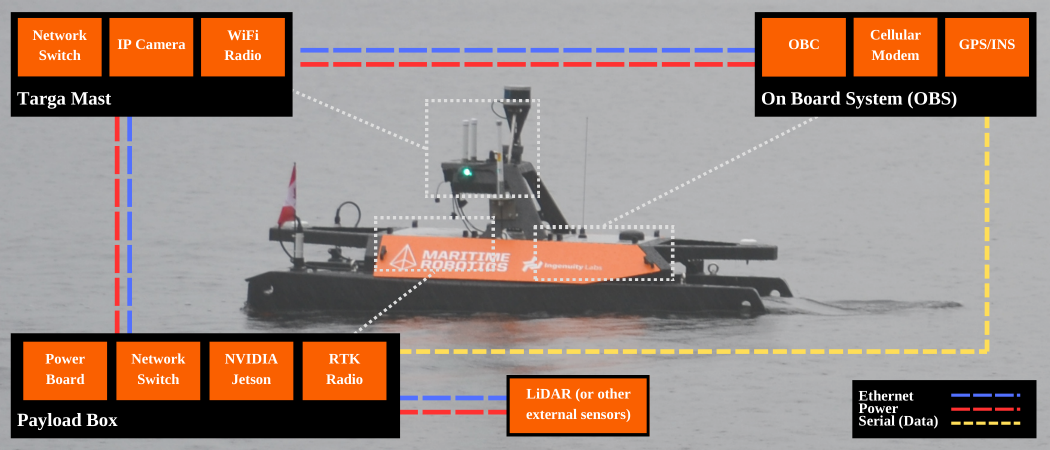}
    \caption{The Otter uses a compartmental design to house electrical systems. Connections between compartments must be made with waterproof connectors and cables. Locations of each compartment indicated in grey.}
    \label{fig:otter_architecture}
\end{figure*}

\subsection{Mechanical Design}

The Otter uses a modular design, with separate waterproof boxes housing different subsystems. A typical Otter will have an ``OBS'' and ``Targa'' box, with the ``Payload'' box reserved for hydrographic instruments. We purchased this box empty to create our own robotics payloads. The breakdown of compartment contents and locations is shown in Fig.~\ref{fig:otter_architecture}.

Commercial off-the-shelf components can be used inside the Otter's sealed compartments. This reduces the need to purchase IP67-rated parts or fabricate individual housings for each component. Connecting the compartments is more difficult because it requires IP67-rated connectors and cables. Sourcing waterproof cables can be a costly and slow process, which other USVs avoid by having a single sealed compartment. To minimize a need for many exterior cables, power and data hubs are used inside the payload box in order to use a single external power and Ethernet cable connection.

\subsection{Power and Data Interfaces}

The Otter provides a daisy chain connection for power and data to an external payload box as illustrated in Fig.~\ref{fig:otter_architecture}. The Ethernet connects other devices to the Otter local area network. The power connection is an unregulated DC supply, so to simplify payload integration, Maritime Robotics sells a regulator board that provides 5 V, 12 V, 30 V, and variable voltage outputs.

Communication between compartments and devices on the Otter is through a local area network.
USV-to-shore communications is with a high-power, directional Wi-Fi system. New devices can be easily added to the network and accessed remotely, making ROS control simple.

The backseat data is transmitted as custom NMEA messages, encoded according to Maritime Robotics specifications. From the Otter, two regularly transmitted messages contain position (latitude, longitude), speed, course-over-ground, orientation, and angular velocity. Additional messages indicate the vehicle mode, motor state, battery capacity, and current power consumption. These data are broadcast over the network at a rate determined by the Otter OBC.

Input commands are similarly encoded as NMEA messages to set the control mode and settings for the vessel. Maritime Robotics provides a number of command options: \textit{drift}, where the motors are turned off, \textit{manual}, where the motor inputs are provided directly, and high level control, such as \textit{course and speed} and GNSS-based \textit{station keeping}.

Although the Otter provides a motor control interface and some navigation data, there are hardware and software system limits. Drawbacks to the Otter backseat driver include:
\begin{itemize}
    \item no accelerometer data from the navigation system;
    \item no position uncertainty or GPS state information;
    \item unsigned motor RPM values (i.e., cannot see which direction propellers are spinning); and
    \item limited sample rate (1--20 Hz) for incoming data.
\end{itemize}
Despite these drawbacks, many of which could be addressed in future software updates from Maritime Robotics, the vehicle has been successfully used in numerous field experiments.

\section{OtterROS}

To use ROS 2 with the Otter USV, a conversion from the OBC backseat driver interface to ROS was required. \textit{OtterROS} enables this communication through ROS 2 topics. New autonomy and sensing algorithms can be built on top of OtterROS to take advantage of the wide range of ROS 2 compatible hardware and software. This section presents an overview of OtterROS and provides an example of how it can be used. For complete details, see the OtterROS documentation and code \cite{otterros_github}.

\subsection{Implementation}

OtterROS uses a publisher-subscriber architecture to let users communicate with the OBC through ROS 2 topics. Lower-level communication between the OBC and OtterROS is encoded as NMEA messages through UDP, so a custom version of {\small\texttt{pynmeagps}} \cite{pynmeagps} is included as a translator for the OtterROS package. New programs can focus entirely on the ROS 2 topics.

A base launch file provides an easy way of starting OtterROS. Additional parameters provide optional connectivity for Velodyne Puck LiDAR and SBG Ellipse-D sensors, and for bagging (i.e., data recording). Bagging is saved as an \textit{mcap} format, which allows easy data sharing and replay on machines that do not have OtterROS.

\subsection{Available Data (Publisher)}

OtterROS publishes all available data from the OBC to the topics listed in Table~\ref{tab:pub_node}. Topics are updated asynchronously as data are received by OtterROS (the actual rate is determined by settings in the Otter OBC). Standard position and orientation topics allow for quick use in ROS 2 visualization tools or in third-party packages.

\begin{table}
    \centering
    \def\arraystretch{1.3}
    \caption{Topics and data published by OtterROS. Standard ROS 2 message types are used when possible.}
    \label{tab:pub_node}
    \input{pub_table}
\end{table}

\subsection{External Commands (Subscriber)}

Commands to the Otter are sent through ROS topics. The OtterROS subscriber is designed to monitor specific command topics and, upon seeing a change, relay the appropriate command to the OBC. It is left to the node publishing to the command topic to determine how often the commands should be sent. Available commands, and their required inputs, are shown in Table~\ref{tab:sub_node}.

\begin{table}
    \centering
    \def\arraystretch{1.3}
    \caption{Topics watched by the subscriber to be sent to the Otter. Commands are relayed to the OBC immediately and only when there is an update to the topic.}
    \label{tab:sub_node}
    \input{sub_table}
\end{table}

\subsection{Development Example: Otter Control}

We developed a Nonlinear Model Predictive Controller (NMPC) for the Otter USV.  This application is briefly presented here as an example of a ROS 2 package built on OtterROS. Complete code is available at \cite{otterros_github} in the {\small\texttt{otter-control}} package. Details about the NMPC implementation are omitted in this paper, but can be found in \cite{riley_thesis}.

\subsubsection{OtterROS Interface}

The NMPC controller subscribes to {\small\texttt{otter\_gps}}, {\small\texttt{otter\_imu}}, and {\small\texttt{otter\_cogsog}} from OtterROS. These messages provide position, orientation, and speed information, respectively. As these messages are published sequentially from OtterROS, the {\small\texttt{message\_filters}} package is used to synchronize near-simultaneous data.

This controller was designed to calculate motor inputs force in order to follow waypoint defined paths, so we use the Otter's \textit{manual} back seat command. We do this by periodically publishing calculated surge and torque values to {\small\texttt{control\_cmds}}. When the OtterROS subscriber sees the change, the command is sent over the network to the OBC.

\subsubsection{Use and Customization}

ROS 2 provides numerous ways to launch multiple nodes. The most convenient way to work with custom nodes and OtterROS is to use the OtterROS launch file {\small\texttt{otter\_launch\_base}}, which starts the core OtterROS publisher and subscriber and begins logging data. Custom nodes can then be run (or additional launch files launched) that interact with OtterROS. The NMPC controller package includes launch file examples that show how to do all of this in a single launch script.

Because the NMPC node is built in Python, it can be quickly modified and updated, and can easily take advantage of external packages (e.g., {\small\texttt{numpy}}). As an example, the controller uses an external nonlinear optimization package (CasADi \cite{casadi}) for real-time optimization.

\subsubsection{Validation}

We tested the NMPC against the manufacturer's built-in waypoint following controller (which, to the best of the authors' knowledge, is a cascaded PI/PD speed/heading controller) on a figure eight path. The NMPC outperformed the supplied controller, demonstrating the utility of OtterROS. We also note that with our current hardware configuration (i.e., the NVIDIA Jetson AGX Orin developers kit), the NMPC ran at 10~Hz with a 4~s horizon.

\section{Otter USV Integration}

Specific hardware may vary by USV and need (e.g., computing platform, sensors), but mechanical and electrical integration with the Otter will be similar for any build. If followed, this section will enable OtterROS to be rapidly deployed on the Otter USV.

\subsection{Computing Platform}

At the heart of OtterROS is a computer platform running ROS 2. We selected the NVIDIA Jetson AGX Orin (developers kit) because it provides the most compute power in this form factor, met the power and space requirements, and is easy to use with USB, Ethernet, Wi-Fi, display, and GPIO interfaces. With growing use of machine learning tools in robotics, the Jetson GPU-focused platform was also preferred over a CPU-focused system like the Intel NUC.

NVIDIA releases their \textit{JetPack SDK} to take advantage of the unique capabilities of the Jetson with ``Jetson Linux''~\cite{nvidia-jetpack}. JetPack 5 is built on Ubuntu 20.04, so OtterROS was developed for ROS 2 \textit{Foxy Fitzroy} (the latest ROS 2 version compatible with Ubuntu 20.04). Users looking to use the latest version of ROS 2 should consider using the JetPack 6 Developers Preview, which is built on Ubuntu 22.04 and compatible with ROS 2 \textit{Humble Hawksbill}.

\subsection{Integration and Supporting Hardware}

\begin{figure}[tb]
    \centering
    \begin{tikzpicture}
        \node[anchor=south west,inner sep=0] (image) at (0,0) { \includegraphics[width=\columnwidth]{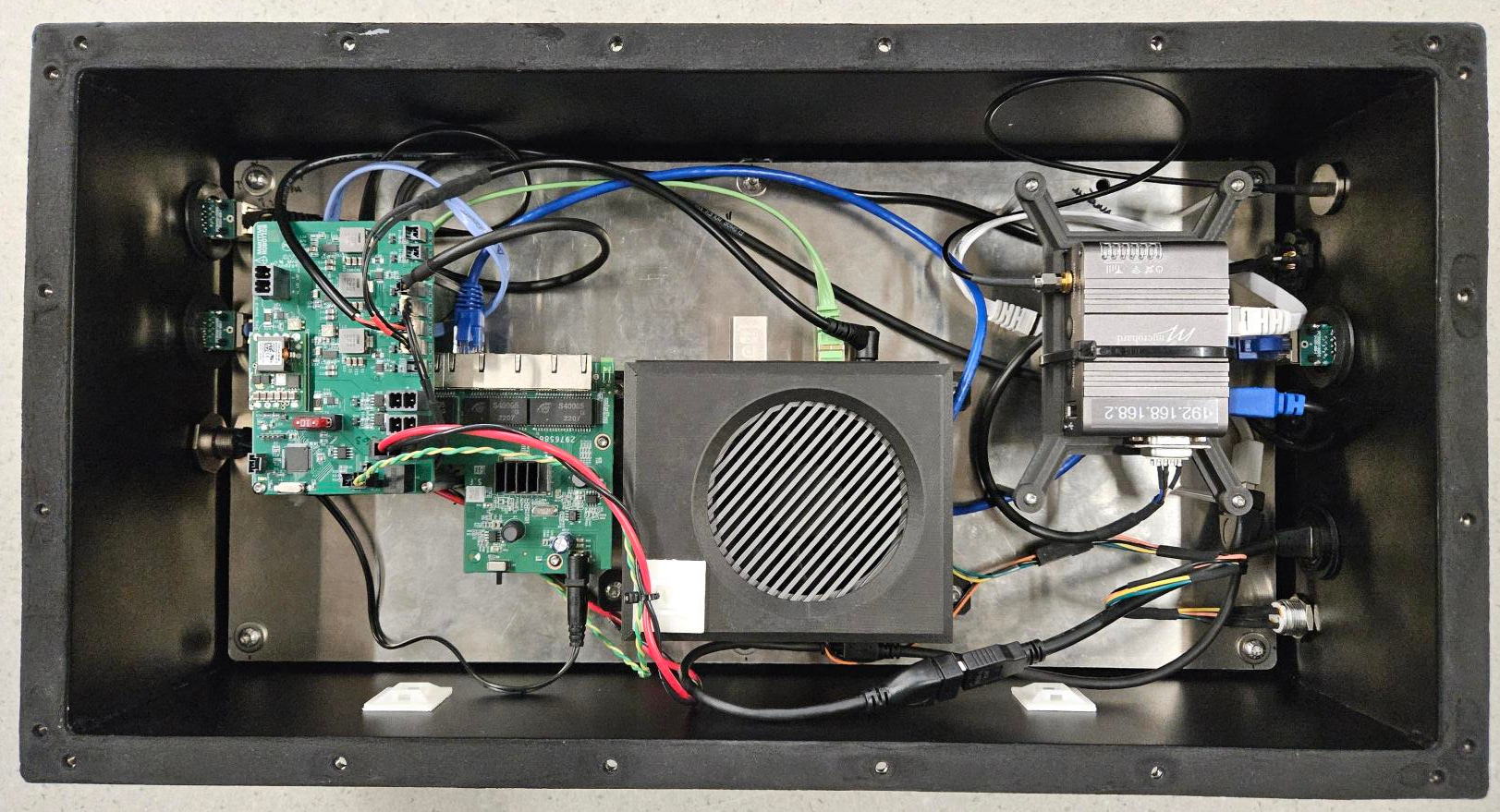} };
        \begin{scope}[x={(image.south east)},y={(image.north west)}]
            \draw[blue,ultra thick,rounded corners] (0.2,0.25) rectangle (0.43,0.6);
            \draw[red,ultra thick,rounded corners] (0.15,0.37) rectangle (0.31,0.75);
            \draw[green,ultra thick,rounded corners] (0.4,0.19) rectangle (0.65,0.58);
            \draw[yellow,ultra thick,rounded corners] (0.70,0.43) rectangle (0.83,0.72);
        \end{scope}
    \end{tikzpicture}
    \caption{Hardware inside the aft (payload) container. Bulkhead connectors are visible on port (left) and starboard (right) walls for connections to other boxes and systems. Main components are the power board (red), network switch (blue), and NVIDIA Jetson inside the mounting cradle (green). An optional RTK radio (yellow) is also shown.}
    \label{fig:payload-integration}
\end{figure}
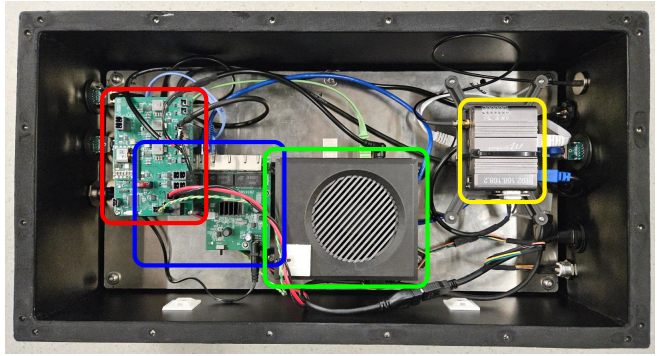

Integrating the Jetson with the Otter USV required additional hardware, cabling, and a mounting solution. The layout of our payload box is shown in Fig.~\ref{fig:payload-integration}, which highlights the most significant components: the Jetson, the Maritime Robotics power board, and a network switch. A list of internal and external parts is presented in Table~\ref{tab:payload_parts}. Full details of our design, including part numbers, drawings, and assembly instructions, can be found at~\cite{offroad_ottermods}.

\begin{table}
    \centering
    \caption{List of components that comprise the Otter payload box for OtterROS operations. Part names and numbers can be found at \cite{offroad_ottermods}.}
    \label{tab:payload_parts}
    \begin{tabularx}{\columnwidth}{
            l
            >{\raggedright\arraybackslash}X
            c}
        \toprule
        \textbf{Description}     & \textbf{Part name}                    & \textbf{Count} \\
        \midrule
        Power distribution       & MR power board                        & 1              \\
        Power input cable        & MR bulkhead power harness             & 1              \\
        Power harness (internal) & 45130-0203 to device                  & 3              \\
        Network switch           & Netgear GS308                         & 1              \\
        Ethernet connectors      & ROP-5SPFFH-TCU7001                    & 3              \\
        Network cables           & Cat6 (generic)                        & 4              \\
        Payload computer         & NVIDIA Jetson AGX Orin Developers Kit & 1              \\
        USB bulkhead             & UA-30PMFL-LC7B20                      & 1              \\
        Switch and indicator     & MPB16-CARE-6-JR-0V                    & 1              \\
        \bottomrule
    \end{tabularx}
\end{table}

The base plate supporting all components is mounted on posts epoxied to the bottom of the box. New plates can be easily swapped for different payloads. This elevated design also protects against any leaks that may form during extended testing, as water can pool underneath the plate without making electrical contact with any components.

When possible, devices were removed from housings and mounted directly to the base plate. The Jetson developers kit is an example where this cannot happen, so a screwless 3D printed ``cradle'' was designed to securely hold it down (pictured in Fig.~\ref{fig:payload-integration}). Metal standoffs were used to maximize use of vertical space in the compartment and allow part placement above the large bulkhead connectors.

\subsection{Optional Hardware}

Additional hardware for RTK corrections and navigation experiments was included in our Otter refit. The Microhard pX2 radio with direct serial connection to the OBS navigation system provides a relay for RTK corrections from a shore-side GPS base station. This device is not IP rated, so is stored inside the payload box. A Velodyne Puck LiDAR was added for navigation experiments and demonstrates how external devices may be added to this design. The Puck is rated IP67, but the power and Ethernet terminations are not, so an adapter box was 3D printed with holes for IP67 connectors. After installing the cables, gaps were sealed with silicone and the box mounted to the underside of the Targa. Hardware designs for both systems are available at~\cite{offroad_ottermods}.

\section{Field Testing Experiences}

We have tested the Otter in a wide range of conditions, across all seasons, and for different robotics experiments. In this section we present lessons learned about USV field work and experiences (both positive and negative) with the Otter USV over two years of testing.

\begin{figure*}
    \centering
    \begin{subfigure}{0.3\textwidth}
        \includegraphics[width=\textwidth]{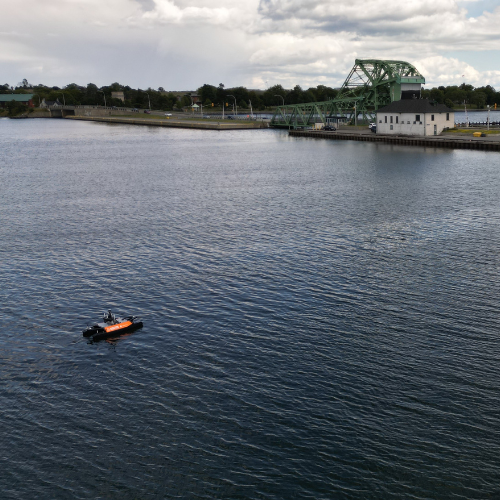}
        \caption{Strong wind and currents cause waves at the base of the Great Cataraqui River.}
        \label{fig:first}
    \end{subfigure}
    \hfill
    \begin{subfigure}{0.3\textwidth}
        \includegraphics[width=\textwidth]{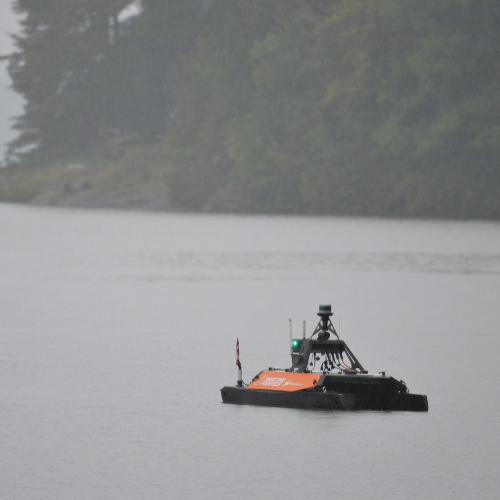}
        \caption{Heavy rain and calm water in Gull Lake (Minden, Ontario, Canada).}
        \label{fig:second}
    \end{subfigure}
    \hfill
    \begin{subfigure}{0.3\textwidth}
        \includegraphics[width=\textwidth]{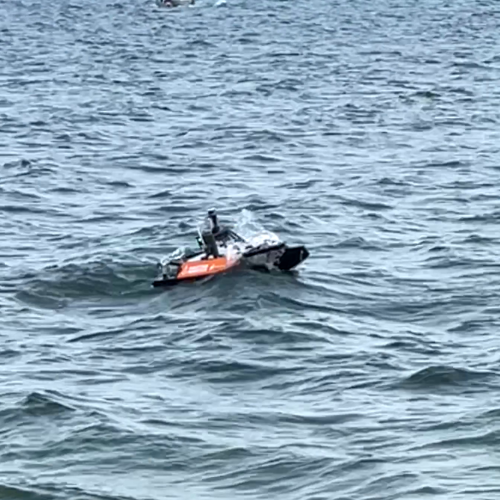}
        \caption{Large waves along the coast of Lake Ontario in Kingston, Ontario.}
        \label{fig:third}
    \end{subfigure}

    \caption{The Otter USV has performed well across a range of environmental conditions.}
    \label{fig:conditions}
\end{figure*}

\subsection{Research Use Cases}

\subsubsection{Wave Mapping}

Initial work with the Otter targeted spatiotemporal mapping of waves~\cite{sears-icra-2023}. This exposed the Otter to waves that were approximately 50~cm high along the coast of Lake Ontario. During these experiments, the Otter was operated with the built-in waypoint-following controller and successfully navigated circular and linear paths. Data from the onboard navigation system was collected through OtterROS by the Jetson for offline processing.

\subsubsection{Model Predictive Control}

Control experiments with the Otter focused on the use of a nonlinear model predictive controller (NMPC) and data driven system modelling to reduce path following errors~\cite{riley_thesis}. Data collection through OtterROS provided the input for offline system modelling. The NMPC used CasADi \cite{casadi} for nonlinear optimization and provided raw motor inputs to the OBC through OtterROS.

\subsubsection{Coastline Localization}

Navigation research using LiDAR scans for coastline matching used OtterROS as the connection between USV navigation data and LiDAR data. By building on ROS 2 and OtterROS, the LiDAR data was easily included by launching existing ROS 2 packages from Velodyne. This work benefited from ROS 2 bagging, which enabled simulated experiments for continued development of the algorithms without a need to return outside.

\subsection{Test Conditions}

The Otter was launched numerous times in Lake Ontario (Kingston, Ontario, Canada), in the Great Cataraqui River (Kingston, Ontario, Canada), and in Gull Lake (Milton, Ontario, Canada). Operating conditions varied, as shown in Fig.~\ref{fig:conditions}, and included:
\begin{itemize}
    \item wave conditions ranging from flat to 50~cm high;
    \item air temperatures from --5 $^\circ$C to 25 $^\circ$C;
    \item heavy rain; and
    \item water temperatures near freezing, ice on the surface.
\end{itemize}

\subsection{System Performance}

In our experiments, the Otter performed well in most conditions, but certain challenges were identified after extensive testing. We also encountered environmental and human factors that affected experiments, which we believe would be common to most USV field experiments. These challenges, and when possible, solutions, are presented in Table~\ref{tab:lessons_learned}.

\begin{table*}
    \centering
    \caption{Summary of USV field testing challenges and experiences.}
    \label{tab:lessons_learned}
    \begin{tabularx}{\textwidth}{
            l
            >{\raggedright\arraybackslash}X
        }
        \toprule
        \textbf{Factor}        & \textbf{Explanation and suggested solution}                                                                                                                                                                                                       \\
        \midrule
        \textit{Otter USV}     &                                                                                                                                                                                                                                                   \\ \midrule
        Motor failure          & Mechanical failure in a motor from unknown cause. Possible propeller strike or due to rapid control commands (to be determined).                                                                                                                  \\
        Motor delay            & The Otter USV has an input delay of approximately 2~s when starting from stationary. A small initial input may mitigate power on delay.                                                                                                           \\
        Communication range    & Difficult to gauge when communications will be lost. Constant monitoring required if Otter is not on a return path.                                                                                                                               \\
        Local network dropouts & In cold conditions, the Otter suffers from network dropouts spanning a few seconds. No solution yet.                                                                                                                                              \\
        \midrule
        \textit{Environmental} &                                                                                                                                                                                                                                                   \\ \midrule
        Motor blockage         & Jammed motors may mean the USV will be unable to return to shore. Scout ahead before deploying boat and seek local experts to avoid any (possible) vegetation.                                                                                    \\
        Ecosystem damage       & USVs can access areas inaccessible by other powered craft, causing damage to mostly undisturbed environments. Researchers should connect with local ecosystem experts to determine if, where, and when it is appropriate for field testing.       \\
        \midrule
        \textit{Human}         &                                                                                                                                                                                                                                                   \\ \midrule
        Legality               & Unlike UAVs, we have not found any rules about USV usage. To avoid any conflict, consult with local authorities (e.g., Coast Guard, government agencies) before starting experiments.                                                             \\
        Runtime                & The Otter has more endurance than most laptop equipment and researchers. Access to power, shelter, and washrooms is critical for a full day of testing (especially in cold conditions).                                                           \\
        Public interaction     & Popular with public, so boaters and swimmers frequently approach the Otter. Limited signage provides no warning to nearby public and no built in audio connection to base station. Requires supervision and patience for interrupted experiments. \\
        Sailing right of way   & When far from shore and encountering another vessel, it is hard to gauge safe passage. Stopping the Otter provides the safest way to let the other vessel pass.                                                                                   \\
        \bottomrule
    \end{tabularx}
\end{table*}

\section{Conclusions}

Options are are limited for robotics researchers looking to work with USVs in ROS 2. To this end, this paper presents a survey of commercial USVs and the process that researchers at Offroad Robotics underwent in selecting a suitable vessel for field robotics research applications. The development of OtterROS, a ROS 2 package for the Otter USV by Maritime Robotics, is discussed and a guide for its integration on the Otter USV is given for researchers looking to leverage these developments. Lessons learned after extensive field testing are shared for researchers active or interested in USV field work, which apply to any USV. With the release of OtterROS and the discussions herein, we hope to help build a growing community of USV robotics researchers.

\section*{Acknowledgments}

Field robotics experiments in Gull Lake (near Minden, Ontario) took place on the traditional territories of the Anishinabee and Mississauga\footnote{\url{https://native-land.ca}\label{native-land}}. Experiments in Kingston, Ontario took place on the traditional territories of the Algonquin, Anishinaabe, Haudenosaunee, and Huron-Wendat\textsuperscript{\ref{native-land}}.

\bibliographystyle{IEEEtran}
\bibliography{references}

\end{document}

%% file: usv-survey.tex

\resizebox{\textwidth}{!}{%
\begin{tabular}{@{}lllrclrrrl@{}}
\toprule
\textbf{USV Name} &
  \textbf{Manufacturer} &
  \textbf{\begin{tabular}[c]{@{}c@{}}Dimensions \\ L$\times$W$\times$H {[}m{]}\end{tabular}} &
  \textbf{\begin{tabular}[c]{@{}c@{}}Draft\\ {[}m{]}\end{tabular}} &
  \textbf{\begin{tabular}[c]{@{}c@{}}Relative\\ Price\end{tabular}} &
  \textbf{\begin{tabular}[c]{@{}c@{}}Software\\ Interface\end{tabular}} &
  \textbf{\begin{tabular}[c]{@{}c@{}}Endurance\\ {[}hours{]}\end{tabular}} &
  \textbf{\begin{tabular}[c]{@{}c@{}}Top Speed\\ {[}m/s{]}\end{tabular}} &
  \textbf{\begin{tabular}[c]{@{}c@{}}Payload\\ {[}kg{]}\end{tabular}} &
  \textbf{Configuration} \\ \midrule
Heron (\textit{discontinued}) &
  \begin{tabular}[c]{@{}l@{}}Clearpath Robotics \\ (Canada)\end{tabular} &
  1.4$\times$1.0$\times$0.3 &
  0.12 &
  \$ &
  \begin{tabular}[c]{@{}l@{}}ROS \\ MOOS-IvP \\ C++ \end{tabular} &
  2 &
  1.7 &
  10 &
  \begin{tabular}[c]{@{}l@{}}S: Differential Drive\\ P: Water Jet\\ H: Catamaran\end{tabular} \\  \midrule
Otter &
  \begin{tabular}[c]{@{}l@{}}Maritime Robotics\\ (Norway)\end{tabular} &
  2.0$\times$1.1$\times$1.1 &
  0.32 &
  \$\$ &
  Backseat Driver &
  20 &
  3.0 &
  30 &
  \begin{tabular}[c]{@{}l@{}}S: Differential Drive\\ P: Propellers\\ H: Catamaran\end{tabular} \\  \midrule
DataXplorer &
  \begin{tabular}[c]{@{}l@{}}Open Ocean Robotics \\ (Canada)\end{tabular} &
  3.6$\times$0.9$\times$1.6 &
  0.5 &
  -- &
  -- &
  \textgreater{}24 &
  3.0 &
  60 &
  \begin{tabular}[c]{@{}l@{}}S: Rudder\\ P: Propellor\\ H: Monohull\end{tabular} \\  \midrule
SKIPPER &
  \begin{tabular}[c]{@{}l@{}}Independent Robotics \\ (Canada)\end{tabular} &
  1.2$\times$0.9$\times$1.0 &
  0.3 &
  \$ &
  ROS2 &
  6 &
  1.0 &
  25 &
  \begin{tabular}[c]{@{}l@{}}S: Differential Drive\\ P: Propellers\\ H: Catamaran\end{tabular} \\  \midrule
Z-Boat 1800-RP &
  \begin{tabular}[c]{@{}l@{}}Teledyne Marine \\ (USA)\end{tabular} &
  1.8$\times$1.0$\times$1.1 &
  0.28 &
  \$\$ &
  Backseat Driver &
  4.5 &
  5.0 &
  30 &
  \begin{tabular}[c]{@{}l@{}}S: Differential Drive\\ P: Propellers\\ H: Monohull\end{tabular} \\  \midrule
WAM V-8 &
  \begin{tabular}[c]{@{}l@{}}Marine Advanced \\ Robotics (USA)\end{tabular} &
  2.5$\times$1.2$\times$1.0 &
  0.1 &
  \$\$\$ &
  \begin{tabular}[c]{@{}l@{}}ROS \\ MOOS-IvP \end{tabular} &
  10 &
  1.5 &
  45 &
  \begin{tabular}[c]{@{}l@{}}S: Differential Drive\\ P: Propellers\\ H: Catamaran\end{tabular} \\ \midrule
SR-Surveyor M1.8 &
  \begin{tabular}[c]{@{}l@{}}SeaRobotics Corporation\\ (USA)\end{tabular} &
  1.8$\times$0.9$\times$-- &
  0.2 &
  -- &
  -- &
  7 &
  2.1 &
   &
  \begin{tabular}[c]{@{}l@{}}S: Differential Drive\\ P: Propellers\\ H: Catamaran\end{tabular} \\  \midrule
SR-Utility 2.5 &
  \begin{tabular}[c]{@{}l@{}}SeaRobotics Corporation\\ (USA)\end{tabular} &
  2.5$\times$1.2$\times$-- &
  0.1 &
  -- &
  -- &
  20 &
  3.8 &
  60 &
  \begin{tabular}[c]{@{}l@{}}S: Differential Drive\\ P: Propellers\\ H: Catamaran\end{tabular} \\  \midrule
Blue Boat &
  \begin{tabular}[c]{@{}l@{}}BlueRobotics\\ (USA)\end{tabular} &
  1.2$\times$0.9$\times$0.5 &
  -- &
  \textless \$ &
  \begin{tabular}[c]{@{}l@{}}Python \\ C++ \\ Rust \end{tabular} &
  \textgreater{}24 &
  3.0 &
  15 &
  \begin{tabular}[c]{@{}l@{}}S: Differential Drive\\ P: Propellers\\ H: Catamaran\end{tabular} \\ \bottomrule
\end{tabular}%
}

%% file: usv-survey-figure.tex
\begin{subfigure}{0.25\linewidth}
    \includegraphics[width=\linewidth]{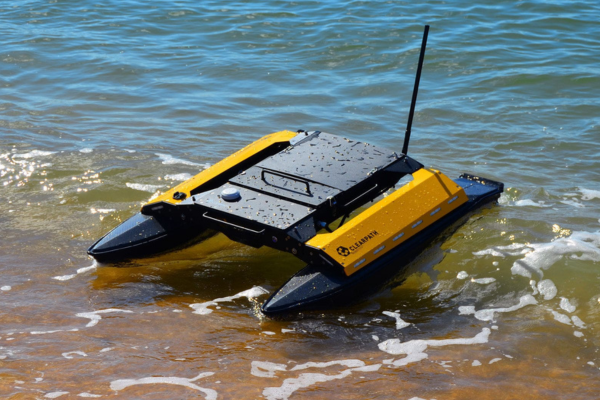}
    \caption{Heron USV \cite{heron}}
    \label{fig:sub_Heron}
\end{subfigure}
\begin{subfigure}{0.25\linewidth}
    \includegraphics[width=\linewidth]{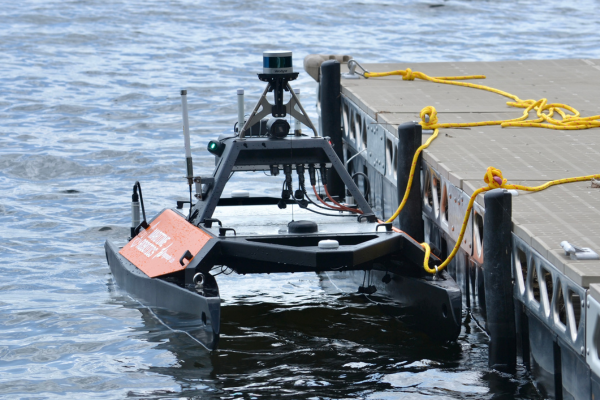}
    \caption{Otter}
    \label{fig:sub_Otter}
\end{subfigure}
\begin{subfigure}{0.25\linewidth}
    \includegraphics[width=\linewidth]{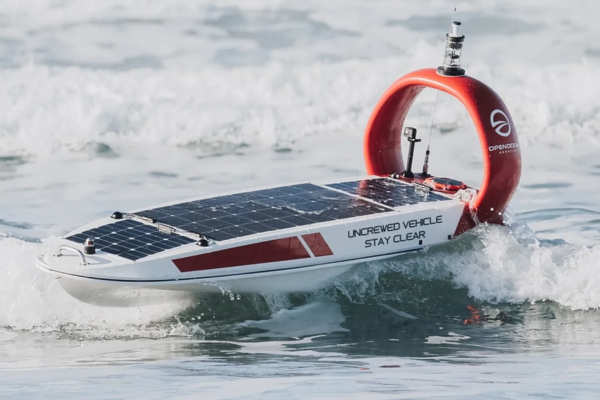}
    \caption{DataXplorer \cite{dataxplorer}}
    \label{fig:sub_DataXplorer}
\end{subfigure}

\vspace{0.25cm}

\begin{subfigure}{0.25\linewidth}
    \includegraphics[width=\linewidth]{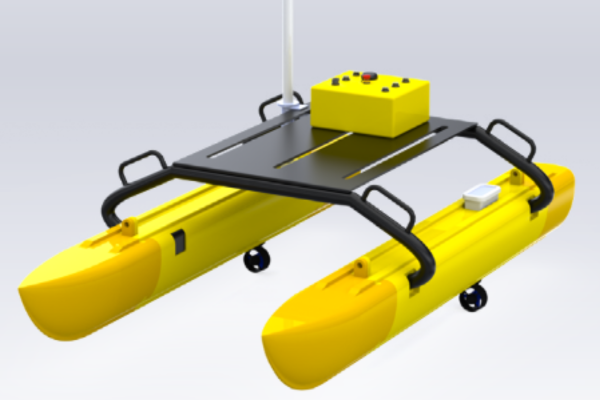}
    \caption{Skipper \cite{skipper}}
    \label{fig:sub_Skipper}
\end{subfigure}
\begin{subfigure}{0.25\linewidth}
    \includegraphics[width=\linewidth]{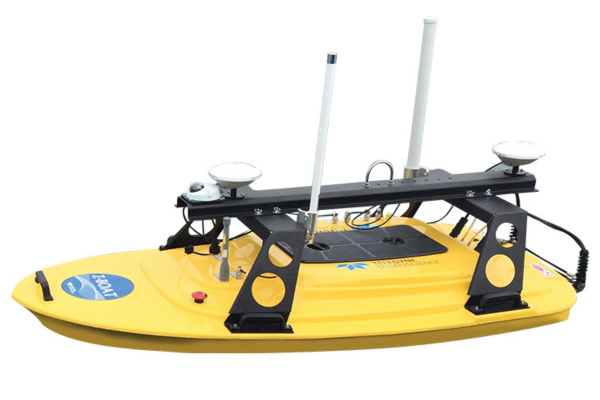}
    \caption{Z-Boat 1800-RP \cite{zboat}}
    \label{fig:sub_zboat}
\end{subfigure}
\begin{subfigure}{0.25\linewidth}
    \includegraphics[width=\linewidth]{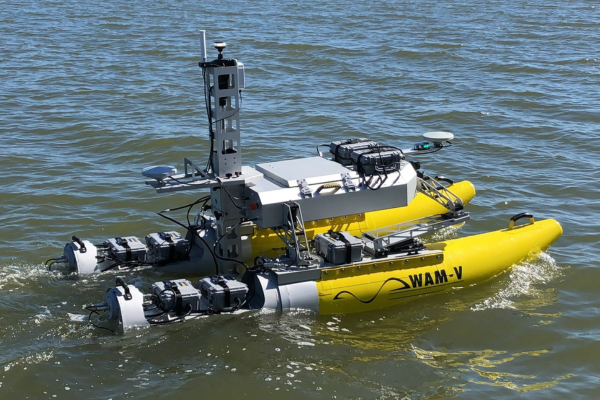}
    \caption{WAM V-8 \cite{wamv8}}
    \label{fig:sub_wamv8}
\end{subfigure}

\vspace{0.25cm}

\begin{subfigure}{0.25\linewidth}
    \includegraphics[width=\linewidth]{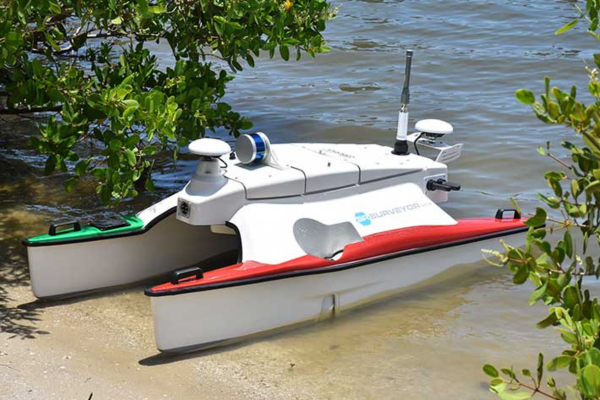}
    \caption{SR-Surveyor M1.8 \cite{sr-boats}}
    \label{fig:sub_surveyor}
\end{subfigure}
\begin{subfigure}{0.25\linewidth}
    \includegraphics[width=\linewidth]{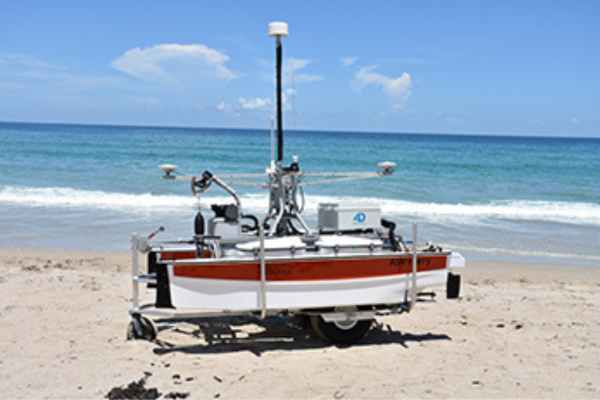}
    \caption{SR-Utility 2.5 \cite{sr-boats}}
    \label{fig:sub_utility}
\end{subfigure}
\begin{subfigure}{0.25\linewidth}
    \includegraphics[width=\linewidth]{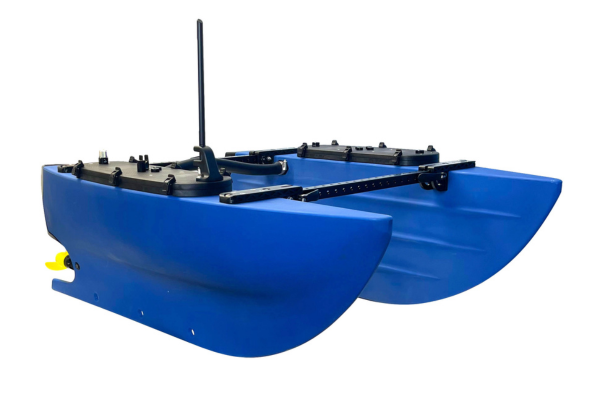}
    \caption{Blue Boat \cite{blueboat}}
    \label{fig:sub_blue}
\end{subfigure}

%% file: usv_factors.tex
    \resizebox{\textwidth}{!}{%
    \begin{tabular}{@{}lll@{}}
        \toprule
        \textbf{Factor} & \textbf{Our (Offroad Robotics) specific considerations} & \textbf{Otter rationale} \\
        \midrule
        \multicolumn{3}{l}{\textit{Mechanical factors}} \\
        \midrule
        Wave conditions & 
            Coastal, 1~m high\footnotemark & 
            Rated for Sea State 2 (0.5~m), so suitable for most days. \\
        Water current &
            St.\ Lawerence River, Great Cataraqui River &
            Max speed greater than anticipated currents for upstream travel \\
        Operation depth &
            Lakes and rivers in Ontario, no hard requirement &
            32~cm draft allows transit through most waterways \\
        Transportation and storage &
            Indoor storage, fits in regular vehicle, launch with two people &
            Fully assembled, fits inside elevator, cargo van, and can be lifted by two people \\
        Payload space &
            Computer and sensing equipment, sonar-ready &
            Large compartments for electronics, designed for hydrographic surveys \\
        Reconfigurability &
            Mounting points, machinable, open surface area & 
            Room for sensors above and below chassis, aluminum structure readily modified \\
        \midrule
        \multicolumn{3}{l}{\textit{Electrical and data factors}} \\
        \midrule
        Runtime &
            Full day testing, overnight experiments &
            20-h rating (at 1 m$/$s), swappable batteries \\
        Power &
            Simultaneous use of computers, active sensors while driving &
            4 kWh provides energy for full-day operations \\
        Remote operations &
            Cover local lakes and rivers, within line of sight &
            0.5--1.0 km maximum range for Wi-Fi connection \\
        OBC interface (physical) &
            Onboard connection (e.g., Ethernet, USB, ...) for data transfer &
            UDP connection over Ethernet for reliable two-way communication \\
        OBC interface (software) &
            ROS 2 (preferred), Python, or similar compatibility &
            No ROS 2 support, but NMEA messages over UDP can be adapted for ROS 2 \\
        System telemetry &
            Navigation, state, system data &
            External interface provides location, speed, and other vehicle data \\
        Remote commands &
            Manual inputs (preferred), speed/heading control &
            External interface allows motor force control and activation of all built-in modes \\
        \midrule
        \multicolumn{3}{l}{\textit{Other factors}} \\
        \midrule
        Availability &
            No export control, shippable to Canada &
            Maritime Robotics has local distributors, can also sell directly \\
        Price &
            Upfront cost, maintenance, operations &
            \$100,000 budget, commercial components readily serviced, minimal consumables \\
        \bottomrule
    \end{tabular}
    }

%% file: pub_table.tex

\begin{tabularx}{\columnwidth}{
>{\raggedright\arraybackslash}m{2.3cm}
>{\raggedright\arraybackslash}m{1.8cm}
>{\raggedright\arraybackslash}X}
\toprule
\textbf{Topic}   & \textbf{Message Type} & \textbf{Populated Parameters}                                         \\ \midrule
\texttt{otter\_gps}       & NavSatFix             & Latitude, longitude, altitude                                                                                                     \\
\texttt{otter\_gps\_time} & OtterTime             & GPS time                                                                                                                              \\
\texttt{otter\_imu}       & Imu                   & Orientation, angular velocity                                                                         \\
\texttt{otter\_status}    & OtterStatus           & Motor RPM, temperature, power consumption
        \\
\texttt{otter\_cogsog}    & COGSOG                & Course, speed over ground, local speed vector                                                                                     \\ \bottomrule
\end{tabularx}

%% file: sub_table.tex

\begin{tabular}{@{}ll@{}}
    \toprule
    \textbf{Command} (\texttt{topic}) 
        & \textbf{Parameters}
    \\ 
    \midrule
    \begin{tabular}[l]{@{}l@{}} Drift
        \\ (\texttt{drift\_cmds}) \end{tabular}
        & On/off flag \\ \midrule
    \begin{tabular}[l]{@{}l@{}} Motor Input (Manual)
        \\ (\texttt{control\_cmds})   \end{tabular}
        & \begin{tabular}[l]{@{}l@{}}Surge force $x \in [-1, 1]$ \\ Sway force $y$ ignored \\Torque $z \in [-1, 1]$ \end{tabular} \\ \midrule
    \begin{tabular}[l]{@{}l@{}} Station Keeping \\
        (\texttt{station\_keeping\_cmds})    \end{tabular}
        & \begin{tabular}[l]{@{}l@{}}NavSatFix (Lat, Lon) \\ Speed $\in \left[0, v_{\text{max}}\right]$ \end{tabular} \\ \midrule
    \begin{tabular}[l]{@{}l@{}}     Course and Speed \\
        (\texttt{course\_speed\_cmds})
   \end{tabular}
        & \begin{tabular}[l]{@{}l@{}}Course $\in \left[0,360\right]$ \\ Speed $\in \left[0, v_{\text{max}}\right]$ \end{tabular} \\
        
    \bottomrule
\end{tabular}